# Automated Defect Identification and Categorization in NDE 4.0 with the Application of Artificial Intelligence

Aditya Sharma*

Department of Mathematics, School of Basic Science, Galgotias University, Greator Noida, India

`aditya.sharma@galgotiasuniversity.edu.in}`

**Abstract.** This investigation attempts to create an automated framework for fault detection and organization for usage in contemporary radiography, as per NDE 4.0. The review's goals are to address the lack of information that is sufficiently explained, learn how to make the most of virtual defect increase, and determine whether the framework is viable by using NDE measurements. As its basic information source, the technique consists of compiling and categorizing 223 CR photographs of airplane welds. Information expansion systems, such as virtual defect increase and standard increase, are used to work on the preparation dataset. A modified U-net model is prepared using the improved data to produce semantic fault division veils. To assess the effectiveness of the model, NDE boundaries such as Case, estimating exactness, and misleading call rate are used. Tiny a90/95 characteristics, which provide strong differentiating evidence of flaws, reveal that the suggested approach achieves exceptional awareness in defect detection. Considering a 90/95, size error, and fake call rate in the weld area, the consolidated expansion approach clearly wins. Due to the framework's fast derivation speed, large images can be broken down efficiently and quickly. Professional controllers evaluate the transmitted system in the field and believe that it has a guarantee as a support device in the testing cycle, irrespective of particular equipment cut-off points and programming resemblance.

## 1      Introduction

Non-Destructive Evaluation (NDE) 4.0 is a major advancement over existing NDE approaches in the fields of industrial manufacturing and infrastructure maintenance [1]. To improve the precision and effectiveness of defect

**\*Corrosponding Author**

identification and classification, this breakthrough combines state-of-the-art technologies, most notably Artificial Intelligence (AI) and Machine Learning

(ML). In addition to revolutionizing defect detection, the integration of AI and ML into NDE operations is expected to improve quality control and predictive maintenance capabilities [2].

With the use of artificial intelligence (AI), NDE 4.0 analyzes enormous volumes of data gathered from several sensors and imaging devices [3]. Traditional NDE techniques, such eddy current, radiography, and ultrasonic testing, mostly rely on human interpretation of results, which can be laborious and prone to subjectivity errors. On the other hand, AI-driven systems make use of complex algorithms that process and analyze data remarkably fast and precisely. Training on large datasets of known flaws, machine learning algorithms are able to detect patterns and abnormalities that might be invisible to the human eye [4]. With the use of this feature, defects may be automatically detected, greatly cutting down on inspection time and improving result reliability.

By giving systems the capacity to learn from data and get better over time, machine learning, a subset of artificial intelligence, is essential to the advancement of NDE 4.0. Annotated datasets with categorized and labelled faults can be used to train machine learning algorithms through the use of supervised learning approaches [5]. The system's predictive abilities improve with each new set of data, resulting in more precise defect definition and classification. Furthermore, data can have hidden patterns found by unsupervised learning techniques that disclose hitherto unidentified kinds of flaws or deviations that might not have been expected.

In NDE 4.0, AI and ML are used to go beyond defect detection and include thorough defect classification. By classifying flaws according to their nature, severity, and possible impact, artificial intelligence (AI) systems can help with maintenance and repair decision-making [6]. Prioritizing interventions, allocating resources optimally, and reducing risks all depend on this classification. In addition, AI-powered predictive analytics may anticipate probable future flaws based on usage trends and historical data, enabling proactive maintenance plans that reduce downtime and increase the lifespan of vital assets [7].

A revolutionary change in defect identification and classification is brought about by the merging of AI and ML with NDE technologies [8]. The overall quality and dependability of industrial and infrastructure systems are improved by NDE 4.0, which takes advantage of these developments to provide more precise, effective, and actionable insights [9]. It is anticipated that these technologies will have an increasing influence on NDE methods as they develop, spurring additional advancements and enhancements in maintenance and defect management techniques.

**1.1 Research Objectives**

- To provide an automated system for industrial radiography fault identification and categorization within the framework of NDE 4.0 by utilizing machine learning and artificial intelligence methodologies.
- To show how cutting-edge semantic segmentation networks can effectively identify weld faults in aerospace components using sparse annotated data, hence addressing the issue of data scarcity.
- To Examine the advantages of virtual flaw augmentation on various dataset sizes in order to improve the defect detection and classification system's performance.

## 2. Literature Review

Naddaf-Sh et al. (2021) [10] give a thorough analysis of the integration of deep learning with digital radiography for the identification and categorization of welding defects. Their work demonstrates how radiography images of welded joints may be analyzed using convolutional neural networks (CNNs). In comparison to conventional techniques, the authors show how deep learning models can greatly improve defect detection's accuracy and dependability. They stress that digital radiography performs better in detecting and categorizing flaws including fractures, porosity, and inclusions when paired with AI approaches. The significance of this study lies in its demonstration of artificial intelligence's ability to automate and enhance the inspection process, hence decreasing human error and enhancing overall safety and quality in welding operations.

Osman, Duan, and Kaftandjian (2021) [11] examine how artificial intelligence can be used in the larger non-destructive evaluation (NDE) setting. Their study offers a thorough review of AI approaches used to several NDE procedures, such as ultrasonic testing, radiography testing, and eddy current testing. It is featured in the Handbook of Non-destructive Evaluation 4.0. The writers talk about how classic NDE methods have been improved by using machine learning algorithms, especially those that deal with pattern identification and anomaly detection. They draw attention to how AI may automate the interpretation of NDE data, increase the rate at which defects are discovered, and shorten the time needed for inspections. Understanding the influence of AI on the sector and how it integrates with current NDE technology is made easier with the help of this chapter.

Papa Georgiou et al. (2021) [12] give an overview of AI technologies for industrial defect identification, emphasizing their usefulness and effectiveness. The writers classify different AI methods, such as supervised and unsupervised learning, and talk about how to use them to find manufacturing problems including dimensional errors and surface imperfections. Their work offers valuable insights into the application of artificial intelligence (AI) in various manufacturing situations, and it was presented at the 2021 International Conference on Information, Intelligence, Systems & Applications. The poll emphasizes the benefits of AI, including its enhanced capacity for flaw detection, affordability, and speedy handling of massive amounts of data. This research advances knowledge about the use of AI to improve quality control in manufacturing processes.

Sun, Ramuhalli, and Jacob (2023) [13] provide a thorough analysis of the uses of machine learning in ultrasonic non-destructive testing (NDE). Their research, which was published in Ultra sonics, focuses on how to improve the interpretation of ultrasonic data for welding flaw detection by utilizing machine learning techniques. The authors examine different machine learning methods, including as supervised and unsupervised learning, and how well they work to improve defect characterization and detection. The review's main conclusions show that machine learning techniques like deep learning and ensemble learning have greatly increased the precision and efficiency of ultrasonic inspections. The research indicates that machine learning (ML) can be a significant improvement over conventional ultrasonic non-destructive testing (NDE) techniques in handling issues including complicated fault patterns and signal noise.

Taheri, Gonzalez Bocanegra, and Taheri (2022) [14] give a thorough examination of smart technologies, AI, and machine learning through non-destructive assessment. Their paper in Sensors explores different AI and ML methods used in NDE, as well as how they might improve and automate the assessment process. The authors discuss a variety of technologies and their uses in various non-destructive evaluation (NDE) techniques, including as eddy current, radiography, and ultrasonic testing, as well as neural networks, support vector machines, and fuzzy logic systems. The review highlights how AI and ML have the potential to revolutionize NDE, especially in terms of enhancing fault detection precision, cutting down on inspection times, and facilitating real-time data processing. Additionally, the integration of smart technologies is covered, emphasizing its significance in the creation of sophisticated, adaptive NDE systems.

Valeske et al. (2022) [15] examine Cognitive Sensor System concepts in the context of NDE 4.0. Their work on integrating AI into sensor systems to improve NDE operations was reported in tm-Technisches Messen. The technology, AI embedding, and the certification and validation of cognitive sensor systems are all covered by the writers. They explain how cognitive sensors with AI algorithms built in may process and interpret sensor data to more effectively identify and categorize faults. The paper also discusses the difficulties of integrating AI into sensor systems, such as problems with system dependability, data quality, and the requirement for thorough validation. This work elucidates the future trajectory of NDE, in which intelligent and cognitive systems are critical to the field's advancement.

## 3. Research Methodology

### 3.1 Data Collection and Annotation

223 CR images that were captured during a certified radiography inspection of airplane welds were used as preliminary data for this investigation. In a total of 3500 cases, defects on the images that could be physically explained included fractures and pores. Additionally, five example plates with artificially created warm fatigue cracks were photographed using a similar radiography device.

**3.2 Data Augmentation**

Techniques for expanding information were applied to improve the preparation set's nature. Randomly divided 512 x 512-pixel patches from the original images were subjected to conventional increasing techniques such as flips, commotion, splendor, arbitrary shear, revolution, cut and resize, and differentiation adjustments. By using relative adjustments, features like weld edges and cracks were kept straight. Explained fundamental flaws were divided and then expanded using the PC's flips, arbitrary noise, and relative changes in order to accomplish virtual imperfection expansion. Then, using physically assigned weld veils, these expanded faults were again inserted into segments that were, in any event, defect-free.

**3.3 Evaluation of Virtual Flaw Data Augmentation**

As anticipated methods for information expansion, standard increase, pure virtual defect expansion, and a blend methodology were all evaluated. The emphasis of pure virtual defect increase was on the use of only improved deficiencies, as opposed to the virtual imperfections used in standard expansion, which were removed haphazardly from the original photographs. Creating a 50% virtual imperfection increase test and a 50% conventional blemish expansion test formed the combined technique. Prior to testing on the test set, the updated informational indexes had to be prepared and approved. A few of the NDE metrics that were used to evaluate the suitability of the various expansion strategies were misleading call rate, probability of detection (POD), and exactness estimation.

**3.4 Deep Learning Model Training**

Using the various upgraded informational collections, a modified U-net model was created to provide defect-versus-foundation semantic division covers. Every model's display was evaluated using the aforementioned NDE metrics. The three expansion techniques were then evaluated using cross-approval, which combined the preparation and approval sets.

**3.5 Deployment and Assessment**

An additional device equipped with a graphics processing unit (GPU) enabled the created deep learning model for problem differentiation. To differentiate between acceptable and undesired flaws, representation devices linked to size, structure, and vicinity models were applied. Master overseers compared the framework's results with those of human investigators to determine whether it was ready for commercial use.

### 3.6 Model architecture

We have decided to use a modified U-net as the model engineering for our automated defect identification and classification framework in light of the analysis conducted by Ranneberger et al. The following components were added to the updated U-net construction:

### 3.6.1 Pre-processing

As a pre-processing step toward working on the meaningfulness of flaws in a manner similar to that of human controllers, unsharp covering was applied to the raw data. In order to detect the initial data, the refined and obscured images were combined into a two-channel input image. The information photos were down sampled from 512x512 to 256x256 in order to lighten the load on the computer.

### 3.6.2 Model Architecture

The modified U-net comprises skip connections between the related organizational phases and is an encoder-decoder network. Using 256x256 pixel picture patches as information sources, a pixelwise characterization cover is generated, with 0 worth pixels addressing the foundation and 1 worth pixel indicating the presence of faults. Less channels are used in the new strategy to speed up the induction and preparation procedures without sacrificing accuracy. The enhanced version of Tensor RT that is used in the organization and the up-sampling method have different qualities, therefore up sampling and convolution activities are used instead of up convolution. Cushioning is used by convolutions to maintain a high and wide gap between pooling layers. The final result is made using a sigmoid enactment capacity for twofold characterisation.

### 3.6.3 Loss Function

A weighted paired cross-entropy misfortune capability is used to represent the uniqueness between all of the foundation and defect pixels. The misfortune capability assigns defective pixels a higher weight than foundation pixels, similar to Ronne berger. For the twofold cross-entropy in this analysis, the hyperparameter weight was set to 3.

### 3.6.4 Training and Optimization

The model is prepared in 32-example increments using the Adam enhancer. The learning rate is a fraction of every 2500 stages, assuming that the approval disaster doesn't abate. Loads from the preparatory step that have the least amount of approval error are reserved for later usage.

**3.6.5 Inference and Post-Processing**

Using a sliding window technique that takes some crossover into account for more notable energy, large images are derived and divided into smaller, more reasonable portions. The picture's line symmetry improves edge flaw detection. After that, the estimated veils are reconnected to form a complete comment cover. To combine neighbouring segments that cross over, use the OR-activity.

Once the full-sized covers are created, recognition measures based on size, shape, and nearness are implemented during post-processing. Markers with thin structures are identified using viewpoint proportions, and approximations are made by fitting square or circle forms around each hidden district. We identify defect groups by proximity, and within these groups we locate chains of porosity. Lastly, the resulting explanations are displayed, with dark circles representing OK flaws and white circles addressing common imperfections.

**Table 1:** Information on train duration, authorization, and testing for every disagreement reaching an agreement on cutting data sets from 100% to 15%: how many shots (7750 × 7750 pixels) were taken for every fold, how many 512 x 512-pixel patches were identified in those images, and how few defects (on average) were discovered outside of the folds in each sector.

| Subset | images | | Unique patches | | Defects (smallest fold) | |
|--------|--------|------------|----------------|------------|-------------------------|------------|
|        | train  | validation | train          | validation | train                   | validation |
| 100%   | 222    | 40         | 141112         | 2823       | 489                     | 387        |
| 75%    | 96     | 33         | 298112         | 2166       | 175                     | 259        |
| 5%     | 48     | 46         | 235112         | 895        | 270                     | 355        |
| 25%    | 18     | 9          | 7512           | 439        | 79                      | 457        |
| 10%    | 23     | 5          | 1365           | 266        | 40                      | 156        |
| 5%     | 6      | 3          | 2149           | 90         | 9                       | 359        |
| 1.5%   | 3      | 3          | 332            | 9          | 5                       | 547        |

## 4. Results

### 4.1 Performance and POD

The intrinsic programmed defect detection and order framework was evaluated using a few models. The models performed well during deduction, as seen by explanation seasons of about 6.3 ms per fix on an Nvidia GTX 3090 graphics card and 15 s on an Nvidia Jetson AGX Xavier running a Tensor RT-changed over model. POD bends are used to show how well a model performed when combined increase was applied to all of the available preparation data. The curves demonstrate how a great deal of imperfection may be seen with remarkable precision. The framework's dependability in identifying flaws was demonstrated by the low a90/95 characteristics that were obtained (a90/95 = fault size perceivable with 90% probability at 95% certainty level).

**Table 2:** The POD architecture promotes a comprehensive learning model constructed using all available training data and the combined (standard and virtual imperfection) expansion.

|  | 0.0 | 0.5 | 1.0 | 1.5 | 2.0 | 2.5 | 3.0 |
|---|---|---|---|---|---|---|---|
| **POD curve 1** | 0.002 | 0.10 | 2.0 | 2.0 | 2.0 | 2.0 | 2.0 |
| **POD curve 2** | 0.004 | 0.03 | 2.0 | 2.0 | 2.0 | 2.0 | 2.0 |
| **POD curve 3** | 0.3 | 0.04 | 2.0 | 2.0 | 2.0 | 2.0 | 2.0 |
| **POD curve 4** | 0.4 | 0.07 | 2.0 | 2.0 | 2.0 | 2.0 | 2.0 |
| **Lower 95% confidence** | 0.03 | 1.0 | 2.0 | 2.0 | 2.0 | 2.0 | 2.0 |

Unit plots for a deep learning model that was created using the complete layout of open information and the two types of expansion (customary and mimicked flaws). For every model that was generated and authorized via the 5-overlay cross-approval method, five case bends are made. Red spots address misses near the base of the plot, and dark specks address hits at its highest point. For every Case bend, the dabbed line corresponds to the absolute bottom of the 95% certainty spans. The lowest certainty limit meeting Unit. a90/95 results in the worst scenario imaginable, as demonstrated by the dabbing even and vertical lines.

### 4.2 Comparison of Data Augmentation Methods

A few metrics were employed to assess the overall advantages of mix growth, unadulterated virtual imperfection increase, and ordinary expansion in terms of advancing information quality.

**Table 3:** The "highest quality level," "unadulterated virtual," and "consolidated" expansion systems are correlated across four NDE personal satisfaction percentage. The outcomes of a five-overlap cross-approval scenario with the worst possible outcomes are displayed. Lower values are beneficial to all actions. (a) 90/95.

|  | **Standard** | **Pure virtual** | **Combined** |
|---|---|---|---|
| **100** | 2 | 3 | 6 |
| **75** | 3 | 5 | 11 |
| **50** | 4 | 7 | 16 |
| **25** | 5 | 9 | 21 |
| **10** | 6 | 11 | 26 |
| **5** | 7 | 13 | 31 |
| **1.5** | 8 | 15 | 36 |

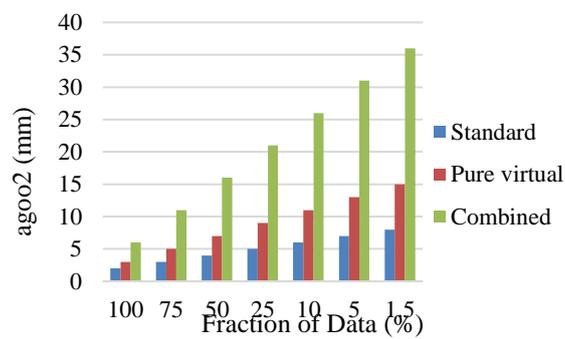

**Figure 1:** Four NDE assessment boundaries and a portion of data are examined for standard, pure virtual, and half-and-half expansion. The scenarios with the highest level of pessimism are displayed using a 5-overlay cross-approval. Lower values are beneficial to all actions. (a) 90/95.

**Table 4:** Across four NDE evaluation borders and a piece of data, standard, pure virtual, and cross-breed rise are examined. From a 5-overlap cross-approval, the most bleak scenario scenarios are displayed. Lower values are advantageous for all actions. (b) A size errors.

|  | **Standard** | **Pure virtual** | **Combined** |
|---|---|---|---|
| **100** | 6 | 3 | 7 |
| **75** | 11 | 5 | 14 |
| **50** | 16 | 7 | 21 |
| **25** | 21 | 9 | 28 |
| **10** | 26 | 11 | 35 |
| **5** | 31 | 13 | 42 |
| **1.5** | 36 | 15 | 49 |

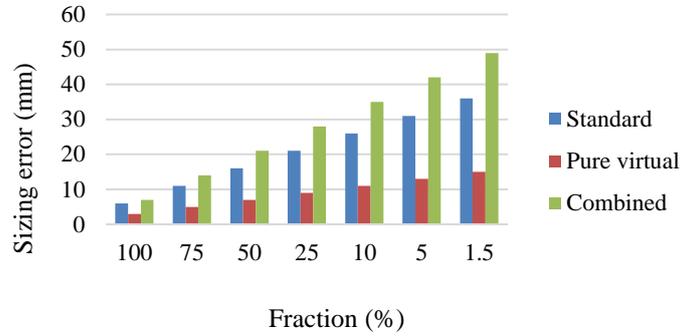

**Figure 2:** Comparison of three growth methodologies' subset data with four NDE evaluation metrics: consolidated, pure virtual, and baseline growth. Based on five-way consensus, the worst-case situation is displayed. A decrease in value is advantageous for all metrics. (b) A size errors.

**Table 5:** The "highest quality level," "unadulterated virtual," and "consolidated" methods of increasing the number of NDEs across four distinct variables are examined. The scenarios with the worst possible five-way cross-approval are displayed. It is preferable to give each activity less weight. (c) A weld area with a false call rate.

|       | Standard | Pure virtual | Combined |
|-------|----------|--------------|----------|
| **100**   | 2        | 6            | 3        |
| **75**    | 3        | 11           | 5        |
| **50**    | 4        | 16           | 7        |
| **25**    | 5        | 21           | 9        |
| **10**    | 6        | 26           | 11       |
| **5**     | 7        | 31           | 13       |
| **1.5**   | 8        | 36           | 15       |

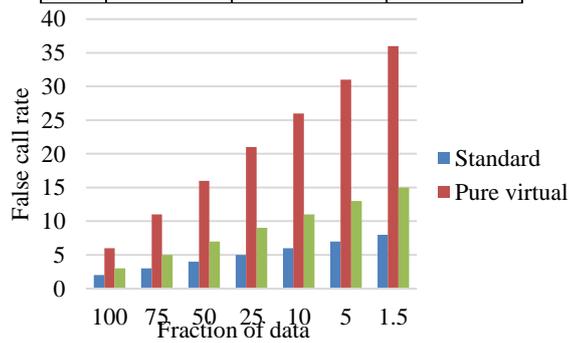

**Figure 3:** Comparison of four NDE assessment measures against incomplete data for three growth strategies (pure virtual, consolidated, and typical). The worst-case

situation is displayed using five-way consensus. Decreasing numbers are good for all measures. (c) A weld area with a false call rate.

**Table 6:** A portion of the data and four NDE assessment factors are used to evaluate standard, pure virtual, and hybrid augmentation. These are the worst-case results of a five-fold cross-validation. Lower numbers are advantageous for all measures. (d) The image's false call rate

|       | Standard | Pure virtual | Combined |
|-------|----------|--------------|----------|
| **100** | 4        | 8            | 10       |
| **75**  | 7        | 15           | 19       |
| **50**  | 10       | 22           | 28       |
| **25**  | 13       | 29           | 37       |
| **10**  | 16       | 36           | 46       |
| **5**   | 19       | 49           | 50       |
| **1.5** | 22       | 50           | 64       |

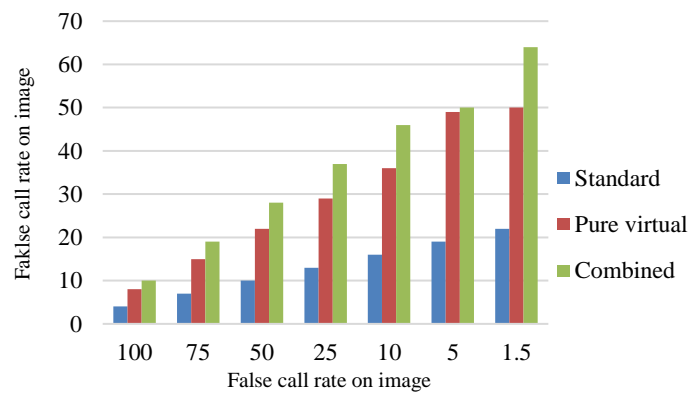

**Figure 4:** The "best quality level," "unadulterated virtual," and "consolidated" increase methods' correlation across four NDE personal satisfaction proportions is shown. Results from a five-crease cross-approval scenario with the worst possible outcome are displayed. Lower values are advantageous for all actions. d) Call rate error on image

- **a90/95:** The results for the worst-case scenario of a 90/95 for determining the size of flaws are displayed in Figure (a). All information subsets showed that the consolidated expansion strategy worked best, suggesting that it is adequate for differentiating faults. The a90/95 attributes remained insignificant in every case when the subset size was reduced, demonstrating the framework's capacity to accurately identify defects of varying sizes.
- **Sizing Error:** The effects of the estimation error are depicted in Figure (b). The presentation that was made using either a blended or standard increase was significant when compared to a 90/95. Once a malformation was properly diagnosed, the framework frequently had the ability to clearly fragment and quantify its magnitude.
- **False Call Rate:** The misleading call rate in the weld area was estimated in order to evaluate the framework's capacity to prevent false alarms. Regardless of how you look at it, the deceptive call rates were roughly 1-2 for every 10 cm of weld, as should be evident in Figure (c). The misleading call rate didn't completely increase even when less than 5% of the preparation information was used.

**4.3 Field Evaluation**

The model that was generated through linked expansion was subsequently presented as a review pipeline feature in an actual test scenario and subjected to quantitative assessment. The radiograph commenting equipment was integrated into a product structure. For the subjective appraisal group, radiography assessment specialists were indispensable. They evaluated the framework's overall value to the review interaction as well as its dependability, transparency, and convenience of use. The findings of the field evaluation should be seen as fascinating rather than authoritative due to the small sample size (three individuals) and equipment and programming requirements.

**5. Conclusıon**

Our investigation got the opportunity to design a robotized framework for flaw identification and arrangement for contemporary radiography within the NDE 4.0 framework. This study addressed the issues brought about by the lack of commented-on data while also demonstrating the advantages of virtual deformity improvement. The framework's low a90/95 attributes achieved in the deformity finding process demonstrate its ability to accurately distinguish flaws of varying sizes. After a thorough evaluation of all available options, combined increase which makes use of both traditional and virtual imperfection expansion was shown to be the most effective technique for increasing information. This discovery emphasizes the value of using more data to improve the profound learning model's suitability. The high derivation speed of the framework allows even very large radiography pictures to be processed quickly. Even though it was small in scope, the field research proved the framework's viability as a roadmap for the review cycle and produced insightful information for ready auditors. The development of systems that can quickly and precisely identify and classify flaws using a variety of inspection techniques is essential to the automated defect identification of NDE 4.0 utilizing AI and ML. Improved material adaptability, real-time predictive analytics integration, and algorithmic refinement to increase accuracy and decrease errors will probably be the major areas of concentration for advancements, which will eventually result in more dependable, efficient, and economical maintenance procedures.